\documentclass[11pt,a4paper]{article}
\usepackage[hyperref]{acl2019}
\usepackage{times}
\usepackage{latexsym}

\usepackage{url}

\usepackage{amsmath}
\usepackage{graphicx}
\usepackage{subcaption}

\aclfinalcopy 
\def\aclpaperid{2431} 

\title{Towards Lossless Encoding of Sentences}

\author{Gabriele Prato\thanks{ ~~Corresponding author: pratogab@mila.quebec}~~~~~~Mathieu Duchesneau~~~~~~Sarath Chandar~~~~~~Alain Tapp\\\\
Mila, Universit\'e de Montr\'eal
}

\date{}

\begin{document}
\maketitle
\begin{abstract}
A lot of work has been done in the field of image compression via machine learning, but not much attention has been given to the compression of natural language. Compressing text into lossless representations while making features easily retrievable is not a trivial task, yet has huge benefits. Most methods designed to produce feature rich sentence embeddings focus solely on performing well on downstream tasks and are unable to properly reconstruct the original sequence from the learned embedding. In this work, we propose a near lossless method for encoding long sequences of texts as well as all of their sub-sequences into feature rich representations\footnote{https://github.com/pratogab/rae}. We test our method on sentiment analysis and show good performance across all sub-sentence and sentence embeddings.
\end{abstract}

\section{Introduction}

Compressing information by encoding it into a fixed size representation in such a way that perfect decoding is possible is challenging. Instead, most of the existing sentence encoding methods focus more on learning encoding such that the encoded representations are good enough for the downstream tasks. In this work, we focus on perfectly decodable encoding of sentences which will be very useful in designing good generative models that can generate longer sentences.


Early efforts such as \citep{Hinton504} have shown autoencoders to effectively yield compressed input representations. \citet{POLLACK199077} was the first to propose using autoencoders recursively. Such models have been shown to be useful for a multitude of tasks. \citet{W13-3512} use recursive neural networks and neural language models to better represent rare words via morphemes. \citet{Socher:2011:DPU:2986459.2986549} use recursive autoencoders for paraphrase detection, learning sentence embeddings \cite{Socher10learningcontinuous} and syntactic parsing. \citet{SocherEtAl2011:RAE} also use a recursive autoencoder to build a tree structure based on error reconstruction. Additionally, \citet{D12-1110} use a matrix-vector RNN to learn semantic relationships present in natural language and show good performance on such task as well as sentiment classification. Then, \citet{sst} introduced the Recursive Neural Tensor Network, trained on a their proposed Sentiment Treebank corpus to better deal with negating sub-sequences for better sentiment classification. Recently, \citet{2017arXiv170101811K} proposed Structural Attention to build syntactic trees and improve even further performance on SST.
Parse trees do alleviate the burden of learning the syntactic structure of text, but these methods limit the number of generated embeddings to the number of nodes in the parse tree. Our proposed method does not have such a restriction as all possible syntactic tree can be simultaneously represented by the architecture.

Convolutional Neural Networks \cite{LeCun:1989:BAH:1351079.1351090} have been used in natural language processing as well. Convolutions work well for extracting low and high level text features and building sequence representations. \citet{AAAI159745} proposed to use CNNs recurrently and show good performance on various language tasks. \citet{NIPS2015_5782,inproceedings} both train CNNs on character level for sentiment analysis, while \citet{2014arXiv1412.1058J} work on word level. \citet{2014arXiv1404.2188K} propose a Dynamic Convolutional Neural Network for semantic modelling of sentences and apply their model to sentiment prediction. Our proposed model is very similar to 1D CNNs. In our case though, we use a multilayer perceptron in parallel instead of a kernel to extract meaningful information out of the layer's input.

Much progress has been made in recent years in the field of general purpose sentence embeddings. Fixed length representations of sentence wide context are learned with the objective of serving for a wide range of downstream tasks. \citet{infersent} trained a bidirectional LSTM on the AllNLI natural language inference corpus \cite{snli,multinli} producing embeddings that generalized well on the SentEval \cite{senteval} benchmark. Following this trend, \citet{GenSen} trained a GRU \cite{gru} on Skip-thought vectors \cite{skipthought}, neural machine translation, parsing and natural language inference to get even better downstream task results. More recently, \citet{BERT,microsoft,2019arXiv190409482L} use Transformers \cite{Transformers} to produce sentence wide context embeddings for each input token and get state-of-the-art results on multiple natural language processing tasks. \citet{2019arXiv190102860D} improve the Transformer method by recursively applying it to fixed length segments of text while using a hidden state to model long dependencies. One downside to these sentence embedding generation methods is that the context is always sequence wide. Our proposed model computes a sentence embedding as well as an embedding for all possible sub-sentences of the sequence with sub-sentence wide context only. All embeddings generated throughout our architecture are constructed the same way and thus share the same properties.

\section{Recursive Autoencoder}
We introduce our recursive autoencoding approach in this section. First we define our model's architecture and how each encoding and decoding recursion is performed. We then describe how the model keeps track of the recursion steps, followed by a description of how the input is represented. We also explain the advantages of using the mean squared error loss for our method. Finally, we dive into the implementation details.

\subsection{Model Architecture}
\begin{figure}[htb]
    \centering
    \includegraphics[width=0.4\textwidth]{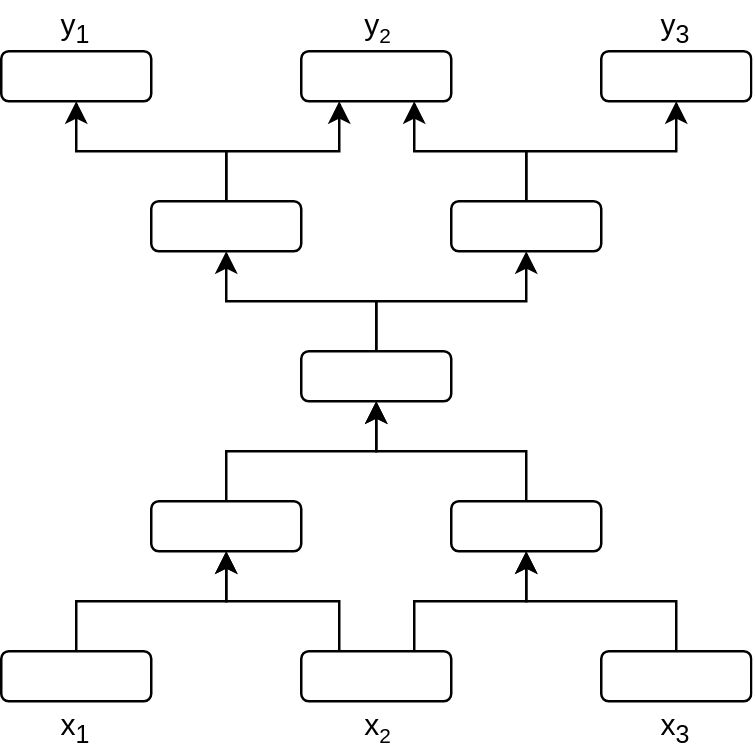}
    \caption{Example of our recursive autoencoder with an input sequence of length three. The encoder recursively takes two embeddings and outputs one until a single one is left and the decoder takes one embedding and outputs two until there are as many as in the original sequence.}
    \label{fig:model_architecture}
\end{figure}

Our model is a recursive auto-encoder. Figure \ref{fig:model_architecture} shows an example of our architecture for a sequence of length three.

The encoder takes an input sequence $\{x_1, \cdots, x_n\}$, where $n$ is the sequence length of the layer's input, and outputs a sequence $\{y_1, \cdots, y_{n-1}\}$. The same $\{y_1, \cdots, y_{n-1}\}$ is then used as input for the next recursion until the output sequence contains only a single element $y_1$, the sentence embedding. The recursion performs the following operation:
\begin{equation}
    y_i = \text{MLP}_{enc}\left(\left[x_i ; x_{i+1}\right]\right) \forall i \in \left\{1, \cdots, n-1\right\}
\end{equation}
where MLP${}_{enc}$ is a shared multilayer perceptron and $\left[x_i;x_{i+1}\right]$ is the concatenation of the embeddings $x_i$ and $x_{i+1}$. MLP${}_{enc}$ is shared throughout all of the encoding recursion steps.

For decoding, it is the inverse procedure of recursively transforming an input sequence $\{x_1, \cdots, x_n\}$ into an output sequence $\{y_1, \cdots, y_{n+1}\}$:
\begin{equation}
    [y_i;y^\prime_{i+1}] = \text{MLP}_{dec}\left(x_i\right) \forall i \in \left\{1, \cdots, n\right\}
\end{equation}
where MLP${}_{dec}$ is the shared multilayer perceptron used by all decoding recursive steps and $[y_i;y^\prime_{i+1}]$ is an embedding twice the size of $x_i$, which we then split into two embeddings $y_i$ and $y^\prime_{i+1}$, each of the same size as $x_i$. Since we obtain two embeddings $y_i$ and $y^\prime_{i+1}$ for each $x_i$, we will have the following embeddings: $y_1$, $\left\{y_2, \cdots, y_n\right\}$, $\left\{y^\prime_2, \cdots, y^\prime_n\right\}$ and $y^\prime_{n+1}$. We merge the overlapping sets by computing the mean:
\begin{equation}
    y_i = \frac{y_i + y^\prime_i}{2} ~\forall i \in \left\{2, \cdots, n\right\}
\end{equation}
and set $y_{n+1} = y^\prime_{n+1}$. We now have a single set of embeddings $\left\{y_1, \cdots, y_{n+1}\right\}$. Both $max$ and $mean$ functions gave similar results, hence we stick with $mean$ throughout all experiments. The output embeddings are then used as input for the next decoding recursion until we get as many elements as the original input sequence.

\subsection{Step Encoding}
To help the recursive autoencoder keep track of the number of recursive steps which were applied to an embedding, we concatenate to the input of MLP${}_{enc}$ the number of the current recursive step as a scalar, starting from 1 for the first recursion, as well as a one-hot of that scalar with custom bucket sizes: \{1, 2, 3-4, 5-7, $\ldots$\}. All buckets after 5-7 are also of size 3. We found this combination of both scalar and one-hot to give best results. When decoding, we also concatenate to the input of MLP${}_{dec}$ this scalar and one-hot, but instead of increasing our recursive step count, we subtract one to it after each recursive decoding step.

\subsection{Input Representation}
We use uncased GloVe embeddings \cite{glove} of size 300 to represent the initial input sequence words, which are then passed through a learned resizing multilayer perceptron (MLP${}_{in}$) before given as input to the encoder. The output of the decoder is also passed through a different learned resizing multilayer perceptron (MLP${}_{out}$) to get back to the GloVe embedding size. We use a vocabulary of 337k words throughout all tasks.

\subsection{Mean Squared Error}
To compute the loss between input GloVe embeddings and the output embeddings, we use the mean squared error (MSE) loss. Obtaining an MSE of 0 would mean our method is lossless, which would not necessarily be the case with the cross entropy loss. MSE also allows us to work with a vocabulary far larger than what is usually the case, as the common classification layer plus cross entropy loss setup tends to have issues with large vocabularies.

\subsection{Implementation Details}
The two embeddings given as input to MLP${}_{enc}$ are each of size $d_{emb}$, as is also its output embedding. Same for MLP${}_{dec}$, the input embedding is of size $d_{emb}$ and the two output embeddings are each of size $d_{emb}$. Both multilayer perceptrons have one hidden layer of size $\frac{3}{2} d_{emb}$, halfway between the input and output size. We apply LayerNorm \cite{layernorm} on the output of each layers of the MLPs, followed by a ReLU activation. The input and output resizing modules MLP${}_{in}$ and MLP${}_{out}$ also have one hidden layer halfway the size of their input and output. They also use ReLU activations, except for MLP${}_{out}$'s last layer. No LayerNorm is used in these resizing components. We test four different $d_{emb}$ embedding sizes in section \ref{ae_experiment}.

\section{Experiments}
In this section, we first present the autoencoding results. Then we present the results on sentiment analysis using our sentence encoding on the Stanford Sentiment Treebank dataset \cite{sst}.

\subsection{Autoencoding} \label{ae_experiment}
As a first experiment, we tested our model on the autoencoding task. Training was done on the BookCorpus \cite{bookcorpus} dataset, comprising eleven thousand books and almost one billion words. At test time, we measured accuracy by computing the MSE distance between an output embedding and the entire vocabulary. We count an output embedding as ``correct" if the closest embedding out of all the vocabulary of size 337k is its corresponding input embedding.

\begin{figure*}[htb]
\centering
\begin{subfigure}{.5\textwidth}
    \centering
    \includegraphics[width=1.0\textwidth]{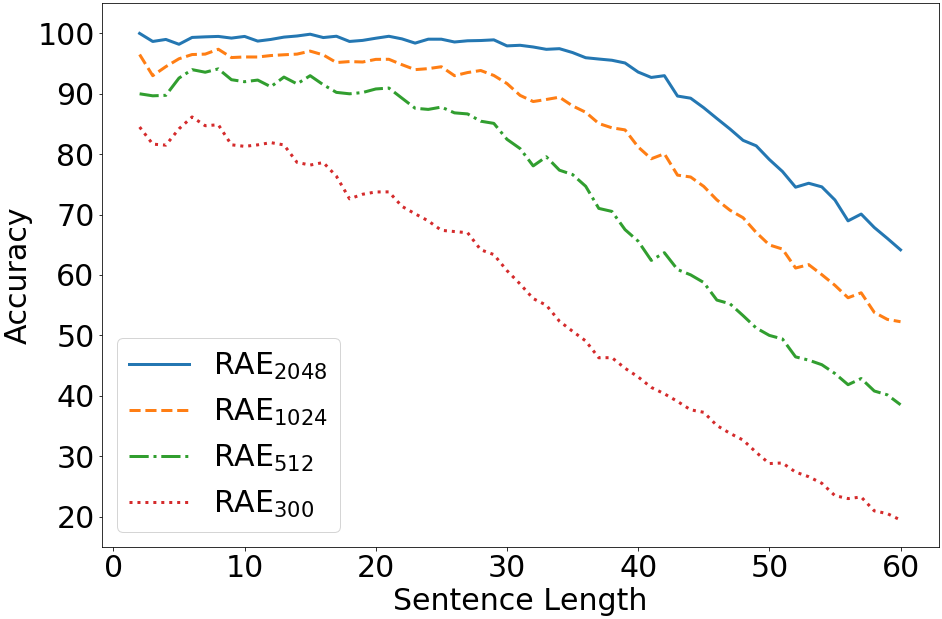}
\end{subfigure}%
\begin{subfigure}{.5\textwidth}
    \centering
    \includegraphics[width=1.0\textwidth]{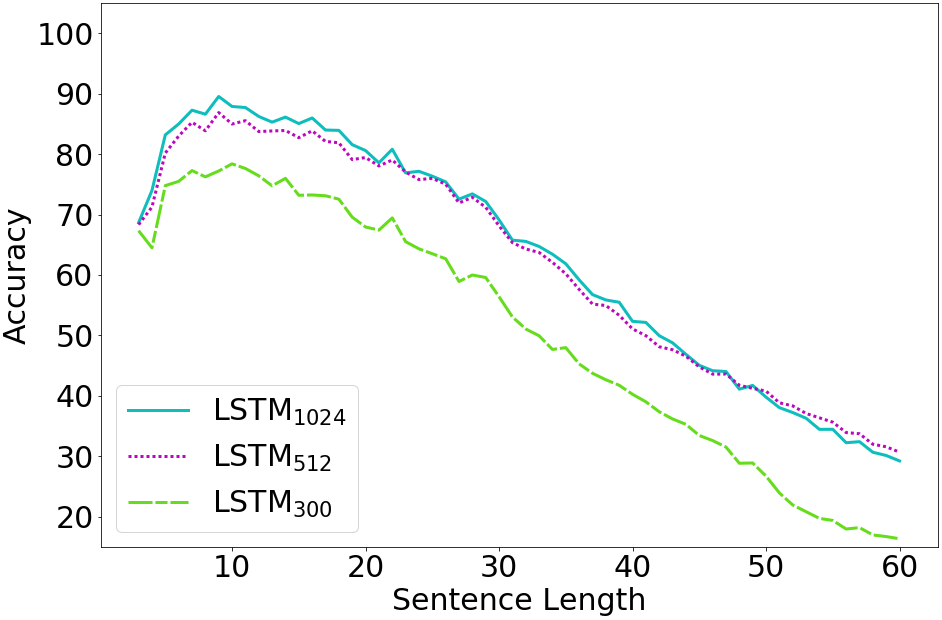}
\end{subfigure}
\caption{Accuracy comparison of different embedding sizes (300, 512, 1024 and 2048) for different sequence lengths. Left is our recursive autoencoder and right a stacked LSTM. An output embedding is counted as correct if the closest embedding out of all the vocabulary is its corresponding input embedding.}
\label{fig:embedding_size}
\end{figure*}

For the autoencoder, we tried four embedding sizes: 300, 512, 1024 and 2048. In all cases, models are given GloVe embeddings of size 300 as input. They also all output embeddings of size 300. Reconstruction accuracy is shown for different sequence lengths in Figure \ref{fig:embedding_size}. With an embedding size of 2048, the model is able to reproduce near perfectly sequences of up to 40 tokens. Longer sentences aren't able to do better and have on average 39 correct tokens. This results in model accuracy linearly going down after a certain threshold, as can be seen in Figure \ref{fig:embedding_size}.

\begin{figure}[htb]
    \centering
    \includegraphics[width=0.5\textwidth]{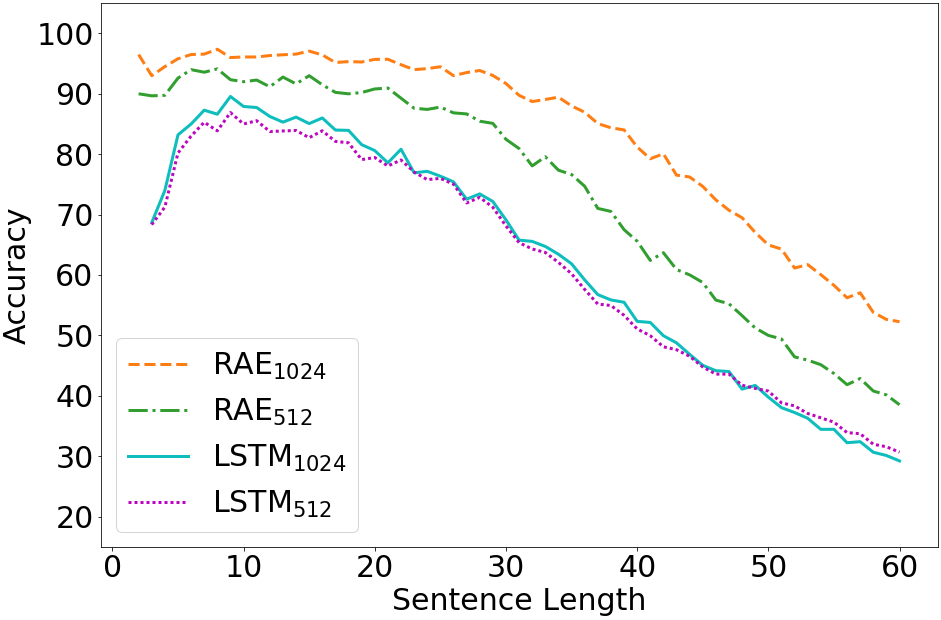}
    \caption{Accuracy comparison of our RAE model versus a stacked LSTM for embedding sizes 512 and 1024. Models of same embedding size have the same capacity.}
    \label{fig:rae_vs_lstm}
\end{figure}

To demonstrate how good our model is at reconstruction, we trained a stacked LSTM on the same autoencoding task. Figure \ref{fig:embedding_size} shows performance of LSTM models for embedding sizes 300, 512 and 1024. All LSTMs have two encoder and two decoder layers. The 1024 variant seems to have reached a saturation point, as it performs similarly to the 512 version. All RAEs and LSTMs were trained for 20 epochs and models with same embedding size have the same capacity. Figure \ref{fig:rae_vs_lstm} shows a better side by side comparison of the RAE and the LSTM for embedding sizes 512 and 1024. Table \ref{tab:mse_losses} shows the MSE loss of all models on the dev set after 20 epochs. The LSTM with an embedding size of 1024 only slightly achieves lower MSE than the RAE with embedding size 300.

\begin{table}[htb]
\centering
\begin{tabular}{lcc}
\hline
Model & $d_{emb}$ & MSE (dev) \\ \hline
LSTM & 300 & 0.0274 \\
& 512 & 0.0231 \\
& 1024 & 0.0191 \\ \hline
RAE & 300 & 0.0208 \\
& 512 & 0.0124 \\
& 1024 & 0.0075 \\
& 2048 & \textbf{0.0019} \\ \hline
\end{tabular}
\caption{Mean squared error loss of stacked LSTMs and our RAE model for different embedding sizes. All models are trained on the autoencoding task for 20 epochs and models of same embedding size have the same capacity. MSE is computed on the BookCorpus dev set \cite{bookcorpus}, between the input GloVe embeddings \cite{glove} and output embeddings.}
\label{tab:mse_losses}
\end{table}

\begin{figure}[htb]
    \centering
    \includegraphics[width=0.5\textwidth]{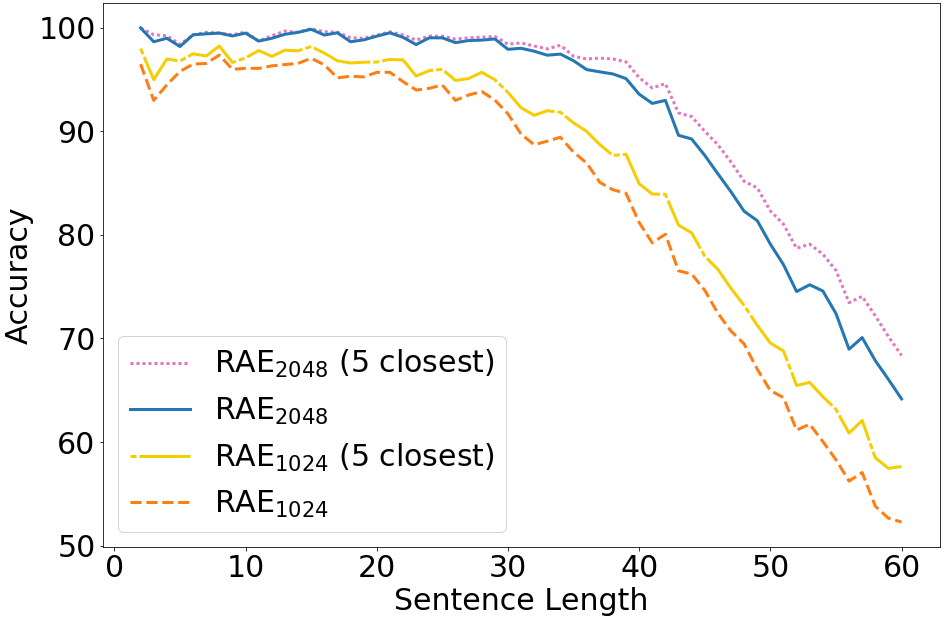}
    \caption{Difference in accuracy when counting an output embedding as correct if the corresponding input embedding is in the five closest versus the closest. Comparison is done on our RAE model with embedding sizes 1024 and 2048.}
    \label{fig:accuracy_range}
\end{figure}

When the output and input embeddings don't match as nearest, they are usually close. Figure \ref{fig:accuracy_range} shows the gain in accuracy for the 1024 and 2048 variants when considering an output embedding as correct if the input embedding is in the five closest to the output, out of all the vocabulary. For the 1024 version, we see on average an increase in accuracy of 2.7\%, while for the 2048 variant, the gain only starts to get noticeable for sequences longer than 30, with an overall average increase of 1.4\%.

\subsection{Sentiment Analysis}
With strong autoencoding performance, one would think that features get deeply encoded into the representation, making it difficult to easily extract them back, which is crucial for a great number of tasks. To this end, we test our architecture on the sentiment analysis task.

The Stanford Sentiment Treebank \cite{sst} is a sentiment classification task where each sample in the dataset is a sentence with its corresponding sentiment tree. Each node in the tree is human annotated, with the leaves representing the sentiment of the words, all the way up to the root node, representing the whole sequence. Comparison is usually done on a binary or five label classification task, ranging from negative to positive. Most models are usually by design only able to classify the root node, while our architecture allows classification of every node in the tree. We use a linear layer on top of each embedding in the encoder to classify sentiment.

We present in Table \ref{tab:sst5_results} results for fine-grained sentiment analysis on all nodes as well as comparison with recent state-of-the-art methods on binary sentiment classification of the root node. For the five class sentiment task, we compare our model with the original Sentiment Treebank results and beat all the models. In order to compare our approach with state-of-the-art methods, we also trained our model on the binary classification task with sole classification of the root node. Other presented models are GenSen \cite{GenSen} and BERT$_{\text{BASE}}$ \cite{BERT}. Both these recent methods perform extremely well on multiple natural language processing tasks. We set the RAE embedding size $d_{emb}$ to 1024. Larger embedding sizes did not improve the accuracy of our model for this task. In this setting, the RAE has 11M parameters, while the models we compare with, GenSen and BERT$_{\text{BASE}}$, have respectively 100M and 110M parameters. Both our model and GenSen fail to beat the RNTN model for the SST-2 task. We see an improvement in accuracy when combining both methods' embeddings, surpassing every model in the SST paper, while being close to BERT$_{\text{BASE}}$'s performance.

Training solely on sentiment classification had same performance as jointly training on the autoencoding task, as the latter had no impact on the sentiment analysis performance. Joint training though had a small impact on reconstruction.

\begin{table}[htb]
\centering
\begin{tabular}{lcc}
\hline
Model & SST-5 (All) & SST-2 (Root) \\ \hline
NB & 67.2 & 81.8 \\
SVM & 64.3 & 79.4 \\
BiNB & 71.0 & 83.1 \\
VecAvg & 73.3 & 80.1 \\
RNN & 79.0 & 82.4 \\
MV-RNN & 78.7 & 82.9 \\
RNTN & 80.7 & 85.4 \\ \hline
RAE & \textbf{81.07} & 83 \\
GenSen & - & 84.5 \\
RAE + GenSen & - & 86.43 \\
BERT$_{\text{BASE}}$ & - & \textbf{93.5} \\ \hline
\end{tabular}
\caption{SST-5 and SST-2 performance on all and root nodes respectively. Model results in the first section are from the Stanford Treebank paper \shortcite{sst}. GenSen and BERT$_{\text{BASE}}$ results are from \cite{GenSen} and \cite{BERT} respectively.}
\label{tab:sst5_results}
\end{table}

\section{Conclusion \& Future Work}
In this paper, we introduced a recursive autoencoder method for generating sentence and sub-sentence representations. Decoding from a single embedding and working with a 337k vocabulary, we manage to get near perfect reconstruction for sequences of up to 40 length and very good reconstruction for longer sequences. Capitalizing on our model's architecture, we showed our method to perform well on sentiment analysis and more precisely its advantage when classifying sentiment trees.

Continuing in the direction of training our model on different NLP tasks, we would like our representations to generalize well on downstream tasks while maintaining their reconstruction property. We would also like to further explore the usage of sub-sentence representations in natural language processing. Finally, we would like to learn our sentence embeddings' latent space, similarly to \citet{NIPS2018_7983}'s method, so as to leverage our autoencoder's strong reconstruction ability and generate very long sequences of text.

\section*{Acknowledgments}
This research was enabled in part by support provided by Compute Canada (\url{www.computecanada.ca}). We would also like to thank Tom Bosc, Sandeep Subramanian, Sai Rajeswar, Chinnadhurai Sankar and Karttikeya Mangalam for their invaluable feedback.

\bibliography{references}
\bibliographystyle{acl_natbib}

\end{document}